\documentclass[letterpaper, 10 pt, conference]{ieeeconf}
\IEEEoverridecommandlockouts
\usepackage{amsmath,amsfonts}
\usepackage{algorithmic}
\usepackage{algorithm}
\usepackage{array}
\usepackage[caption=false,font=normalsize,labelfont=sf,textfont=sf]{subfig}
\usepackage{textcomp}
\usepackage{stfloats}
\usepackage{url}
\usepackage{verbatim}
\usepackage{graphicx}
\usepackage{cite}
\usepackage{graphicx}
\usepackage{amsmath}
\usepackage{amssymb}
\usepackage{booktabs}
\usepackage{multirow}
\usepackage{tabularx}
\usepackage{cite}
\usepackage{color}
\usepackage[pagebackref=false,breaklinks=true,letterpaper=true,colorlinks,bookmarks=false]{hyperref}
\begin{document}

\title{Can Single-View Mesh Reconstruction Generalize to Robot Camera Rotation?}

\author{Yu Zhan$^{1,2}$, Guangcheng Chen$^{1}$, Hanjing Ye$^{1}$, Zhiqin Cheng, Zanjia Tong$^{1}$, 
\\Wenjun Xu$^{2}$, and Hong Zhang$^{1*}$, \textit{Life Fellow, IEEE}% <-this % stops a space
\thanks{*corresponding author (hzhang@sustech.edu.cn).}% <-this % stops a space
\thanks{$^{1}$Yu Zhan, Guangcheng Chen, Hanjing Ye, Zanjia Tong and Hong Zhang are with Shenzhen Key Laboratory of Robotics and Computer Vision, Southern University of Science and Technology (SUSTech). $^{2}$Yu Zhan and Wenjun Xu are with Pengcheng Laboratory. }
}

\maketitle

\begin{abstract}
Single-view mesh reconstruction predicts object meshes and spatial layouts from a single observation, making it attractive for fast robot spatial reasoning and real-to-sim digital twins. However, robot-mounted cameras naturally rotate during manipulation and navigation, while learned single-view reconstruction models often rely on view-dependent priors and may generalize poorly to out-of-distribution camera rotations. Such rotations can introduce 3D inconsistencies, incorrect layouts, and violations of physical constraints, but this failure mode remains under-evaluated. We introduce an evaluation protocol with controlled axis-wise roll, pitch, and yaw sweeps to trace errors in monocular depth estimation (MDE), canonical object meshes, camera-space layout, and physical plausibility within a representative SAM3D-style pipeline. On the Aria Digital Twin dataset and a real Franka wrist-camera sequence, camera rotations induce MDE distortion, layout drift, and collision penetration, while canonical mesh predictions remain relatively stable. A two-stage SAM3D+FoundationPose pipeline is more robust than one-stage feed-forward layout prediction, and our Gravity-Aware Refinement reduces one-stage pairwise ICP-based layout-orientation error by 47.1$\%$. Our evaluation reveals that current single-view mesh reconstruction methods generalize poorly to robot camera rotation, and suggests that explicit gravity cues are important for reliable robotic single-view mesh reconstruction.
\end{abstract}

% \begin{IEEEkeywords}
% Deep Learning for visual perception, object detection, segmentation and categorization, perception for grasping and manipulation
% \end{IEEEkeywords}

\section{INTRODUCTION}
\label{sec:intro}

Can single-view mesh reconstruction generalize to robot camera rotations? This question matters because single-view mesh reconstruction, which predicts the canonical object meshes and spatial layouts of visible objects from a single RGB or RGB-D observation, has advanced rapidly in computer vision and is increasingly attractive in robotics. Fast object-level geometry from a single view can support robot spatial reasoning and the construction of real-to-sim digital twins.~\cite{kapelyukh2026zerobot,agarwal2026scenecomplete,xiang2026clutteredrealtosim}. Traditionally, building such 3D object models required labor-intensive assets, multi-view capture, or dense scene scans~\cite{nolte2025singleviewrobotics}. Recent 3D foundation models trained on large-scale data, such as SAM3D~\cite{sam3d2025}, make a different workflow possible: given a single RGB-D observation, or a single RGB observation paired with a monocular depth estimation (MDE) model, a feed-forward pipeline can predict object-space canonical meshes and their camera-space spatial layout.

\begin{figure}[t]
    \centering
    \includegraphics[width=0.98\linewidth]{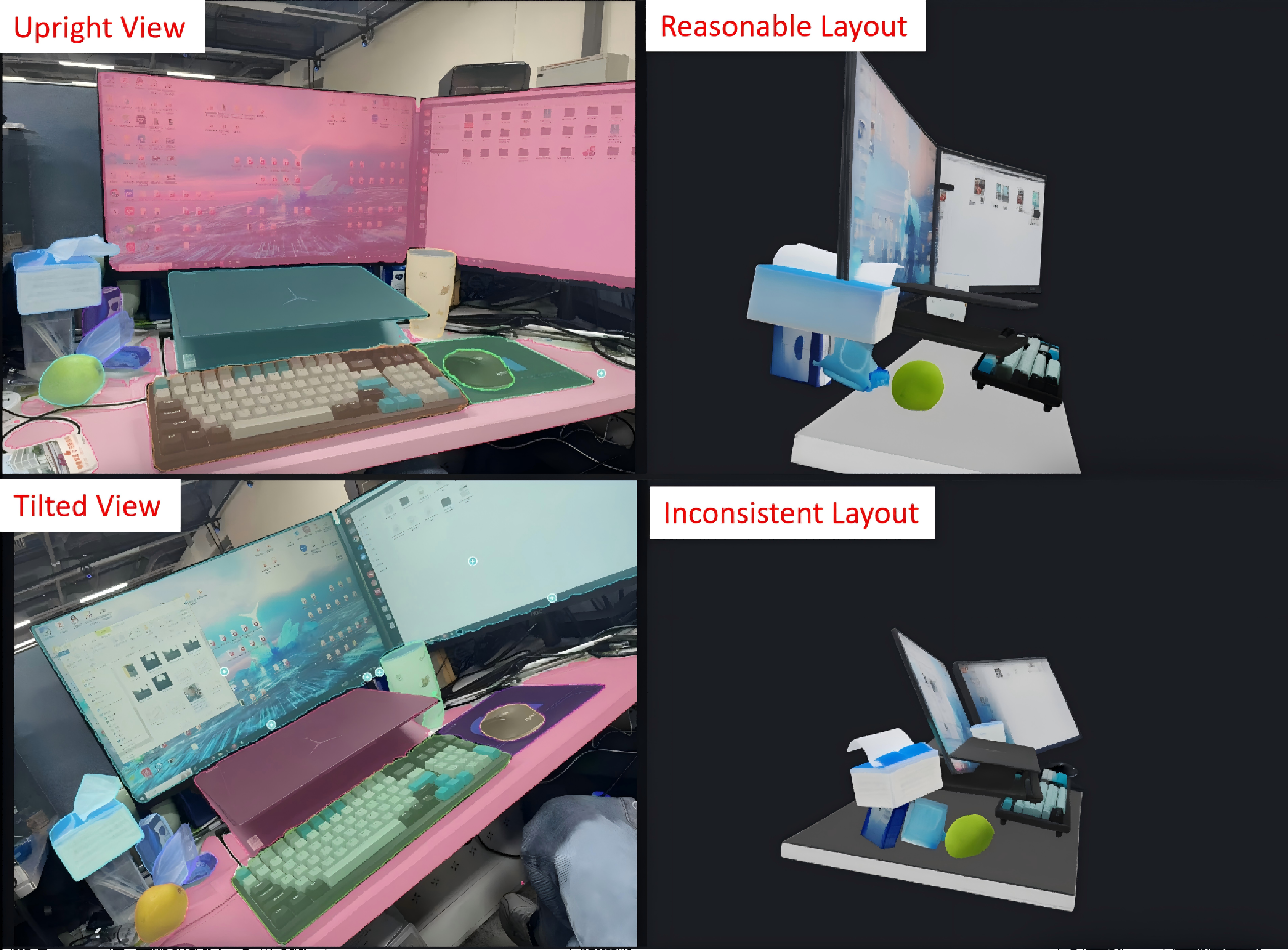}
    \caption{Rotation-induced layout failures. When the upright view yields a reasonable scene layout, the same objects can remain visible after camera rotation but receive inconsistent layout predictions, which in turn trigger mesh penetration and support errors. A model robust to view changes should not exhibit this failure pattern.}
    \label{fig:teaser}
    \vspace{-0.30in}
\end{figure}

However, single-view mesh reconstruction is still not ready to be treated as a reliable robotics primitive. A central weakness is viewpoint generalization, including camera rotation. When the camera observes an object from a rotated view that differs from the dominant training-view distribution, current single-view mesh reconstruction methods face a distribution shift so that their inferred pose and scale may no longer be consistent. Fig.~\ref{fig:teaser} shows this failure mode for SAM3D: after camera rotation, the same scene can yield inconsistent object pose, leading to mesh penetrations that violate physical constraints. Such rotated views are common for cameras mounted on robot wrists, bodies, heads, and aerial platforms \cite{brazil2023omni3d,liang2025pi3det,zhan2025monocular}. The resulting 3D errors and cross-view inconsistencies are harmful to stable perception, collision checking, and robot world modeling. Understanding how single-view mesh reconstruction behaves under robot camera rotation is therefore essential before using these models as robotics primitives.

\begin{figure*}[t]
\vspace{0.1in}
\centering
\includegraphics[width=0.95\linewidth]{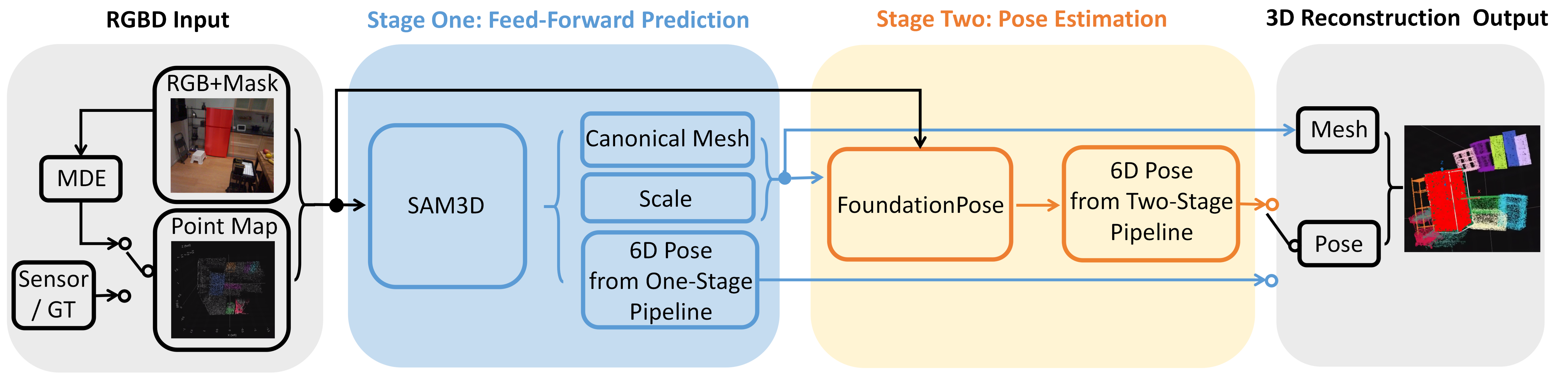}%
\vspace{-0.1in}
\caption{SAM3D pipeline and the two evaluated pipelines. The one-stage pipeline uses SAM3D's feed-forward layout prediction to place canonical object meshes in the camera frame, while the two-stage pipeline reuses the canonical mesh and estimates the downstream object pose with FoundationPose.}
\label{fig:sam3d_pipeline}
\vspace{-0.1in}
\end{figure*}

Current robotics-oriented evaluations and systems have not treated camera rotation as a controlled variable for single-view mesh reconstruction~\cite{nolte2025singleviewrobotics}. Recent robot systems already use single-view mesh reconstruction for manipulation and real-to-sim scene construction~\cite{kapelyukh2026zerobot,agarwal2026scenecomplete}. Other works have noticed that reconstructed layouts may be physically implausible and address this issue through physical constraints~\cite{malenicky2025physpose,yu2026picasso}. No study has systematically quantified how a SAM3D-style model performs when the same scene is observed under controlled camera rotation. This paper fills that gap.

Fig.~\ref{fig:sam3d_pipeline} summarizes the pipeline of our study. SAM3D takes an image, object masks, and an image-aligned point map, and predicts an object-space canonical mesh and a camera-space layout. We compare two ways of turning these predictions into a scene. The one-stage pipeline uses SAM3D's own feed-forward layout branch. The two-stage pipeline reuses the predicted canonical mesh but delegates camera-space pose estimation to FoundationPose~\cite{wen2024foundationpose}, a widely used object pose estimator that registers a known object mesh to an RGB-D observation. This split is useful because it lets us ask where rotation sensitivity enters: the point map, the canonical mesh, or pose and scale.

It remains unclear how monocular scene geometry errors behave as a component inside a full single-view mesh reconstruction pipeline under robot camera rotation. This question is important because the upstream point map may come from either sensor depth or monocular depth estimation (MDE), and learned monocular geometry can be strongly affected by training-view distribution shift under robot camera rotation~\cite{nugent2025pde}. Recent MDE foundation models~\cite{wang2025moge,wang2025vggt,lin2025depthanything3,piccinelli2025unidepthv2,hu2025metric3dv2} achieve great progress, but it is still underexplored how their rotation-dependent geometry errors appear and propagate when used as upstream components in single-view mesh reconstruction. This motivates a controlled rotation-based analysis that traces errors from MDE to object layout and physical plausibility.

We therefore study a representative SAM3D-style pipeline under controlled camera rotation. To enable scalable evaluation, we synthesize rotated egocentric observations from the wide-FOV images of the Aria Digital Twin (ADT) dataset~\cite{pan2023adt} using crop-based optical-center rotations. Under this protocol, we evaluate view consistency using metrics for MDE distortion, object-space canonical mesh stability, camera-space layout consistency, and physical plausibility. Besides SAM3D, we also evaluate four additional single-view mesh reconstruction methods\cite{dogaru2025gen3dsr,huang2025midi,meng2026scenegen,Zhao2025DepR}, and further validate the failure mode with a real Franka wrist-camera rotation sequence. 

Our results trace a clear failure chain. Camera rotation degrades MDE quality; object-level MDE distortions are coupled with layout errors; camera-space layout is substantially more fragile than object-space canonical mesh prediction; two-stage pipelines are more stable than one-stage feed-forward layout prediction. These trends appear not only on ADT but also in the real wrist-camera sweep, suggesting that camera rotation is a practical out-of-distribution stressor for robotic single-view mesh reconstruction. Based on the relative stability of canonical object meshes, we further propose Gravity-Aware Refinement (GAR) that regularizes object orientation using gravity cues to improve the stability of one-stage layout prediction.
Our contributions are threefold:
\begin{itemize}
\item We introduce a rotation-controlled evaluation protocol for scalable view-consistency analysis of single-view mesh reconstruction under camera roll, pitch, and yaw sweeps.
\item We systematically trace the rotation-induced failure chain across MDE point map, object mesh, camera-space layout, and physical plausibility on ADT and a real Franka wrist-camera sequence, while comparing one-stage and two-stage pipelines of SAM3D.
\item We propose Gravity-Aware Refinement (GAR), which leverages the relative stability of canonical object meshes and gravity cues to improve one-stage layout stability under camera rotation.
\end{itemize}

\section{RELATED WORK}
\label{sec:related}

Our focus is on a SAM3D-style single-view mesh reconstruction pipeline. We review the fields most relevant to our question: single-view mesh reconstruction for robot scene hypotheses (Sec. \ref{related:single3D}) and viewpoint sensitivity under robot camera rotation (Sec. \ref{related:sensitivity}).

\subsection{Single-View Mesh Reconstruction for Robotics}
\label{related:single3D}
Single-view mesh reconstruction aims to predict the canonical object meshes and spatial layouts of visible objects from a single RGB or RGB-D observation, enabling fast 3D scene understanding when multi-view capture is unavailable or costly. Recent methods, including Gen3DSR~\cite{dogaru2025gen3dsr}, MIDI~\cite{huang2025midi}, SceneMaker~\cite{shi2025scenemaker}, SceneGen~\cite{meng2026scenegen}, DepR~\cite{Zhao2025DepR}, and SAM3D~\cite{sam3d2025}, have advanced this direction for object and scene reconstruction. In this paper, we study SAM3D because its point-map-conditioned pipeline exposes the full chain from upstream scene geometry to downstream object layout, making it suitable for diagnosing rotation-induced failures. This differs from multi-view object reconstruction methods such as ShapeR~\cite{siddiqui2026shaper}, which rely on posed image sequences; single-view mesh reconstruction instead predicts object shape and layout under limited-view conditions.

Robotics systems use these outputs in both integrated and staged ways. ZeroBot~\cite{kapelyukh2026zerobot} generates a mesh from a single view for rapid real-to-sim learning and uses FoundationPose~\cite{wen2024foundationpose} for downstream tracking. SceneComplete~\cite{agarwal2026scenecomplete} and Xiang et al.~\cite{xiang2026clutteredrealtosim} build staged pipelines in which geometry recovery, registration, and physical reasoning are separate steps; the latter uses SAM3D+ICP as an initialization and notes that SAM3D-predicted geometry and transforms can deviate from the real world. 

Recent pose refinement methods show that physically plausible object layout estimates benefit from explicit structural constraints: PhysPose~\cite{malenicky2025physpose} refines 6D object poses through a post-processing optimization with non-penetration and gravitational constraints, while Picasso~\cite{yu2026picasso} performs holistic scene reconstruction with physics-constrained sampling to reduce inter-object penetration and floating artifacts. Beyond rigid-object settings, DICArt~\cite{zhang2026dicart} advances category-level articulated object pose estimation with hierarchical kinematic coupling under partial and challenging observations. Finally, Nolte et al.~\cite{nolte2025singleviewrobotics} evaluate whether single-view mesh reconstruction models meet robotics requirements such as immediacy, physical fidelity, and simulation readiness. These works establish the robotics motivation and physical-plausibility gap, but they do not isolate camera rotation as a controlled variable for evaluating layout stability.

\subsection{Viewpoint Sensitivity under Robot Camera Rotation}
\label{related:sensitivity}
Learned monocular 3D systems depend strongly on priors from their training distributions, making viewpoint shift a central challenge in robot settings where cameras move with bodies, wrists, heads, or aerial platforms. Although viewpoint sensitivity has not been systematically evaluated for single-view mesh reconstruction, related forms of this issue have been studied in several monocular 3D subfields. For 3D bounding box detection, Omni3D~\cite{brazil2023omni3d} underscores the role of camera and viewpoint diversity by introducing a large-scale cross-dataset benchmark, exposing generalization challenges under varying camera configurations. In robot learning, Jiang et al.~\cite{jiang2025knowyourcamera} show that policies can fail under viewpoint shifts and improve when explicitly conditioned on camera geometry.

In monocular depth estimation (MDE), Camera Pose Matters~\cite{zhao2021cameraposematters} identifies camera-pose distribution bias and proposes perspective-aware augmentation and pose conditioning. PDE~\cite{nugent2025pde} evaluates depth robustness with controlled procedural perturbations and reports camera perturbations as challenging, while How to Evaluate Monocular Depth Estimation~\cite{wu2025evalmde} argues that common depth metrics can miss important geometric changes. 3DRot~\cite{yang20253drot} uses camera-centric pure rotations to create rotated views for RGB-based 3D tasks. However, 3DRot focuses on data augmentation, where changes to the image footprint and canvas padding are acceptable. In contrast, we first use a crop-based homography transformation to generate rotated views for controlled evaluation to measure how camera rotation affects the single-view mesh reconstruction pipeline.

\section{METHODOLOGY}
\label{sec:method}

\subsection{Pipeline Overview}
\label{sec:method_setup}
We study a representative single-view, point-map-conditioned mesh reconstruction pipeline under controlled camera rotation. For a source frame $f$ and a camera rotation $a$ (a pitch, roll, or yaw angle), we denote the rendered observation by $I_f^{(a)}$. As illustrated in Fig.~\ref{fig:method_protocol} (a), the protocol changes camera orientation under a fixed virtual camera. We evaluate two input conditions: a GT point map, which isolates downstream gravity sensitivity, and an MDE point map, which lets us test whether upstream monocular errors are coupled with downstream layout errors.

Fig. \ref{fig:sam3d_pipeline} summarizes the two pipelines we study. Both start from the same inputs: an RGB image, an object mask, and a point map. The one-stage pipeline directly uses SAM3D's feed-forward predictions. For each valid object, SAM3D predicts an object-space canonical mesh and a camera-space layout, which places the mesh back into the scene. We assemble the scene from all valid objects and use them as evaluation units.

The two-stage pipeline keeps the same SAM3D canonical mesh but replaces SAM3D's layout branch with FoundationPose. Specifically, we first scale the canonical mesh using SAM3D's estimated object scale, and then use FoundationPose to estimate the object's camera-space pose. This comparison separates canonical mesh prediction from layout prediction, allowing us to test which component is more sensitive to controlled camera rotation. 

Sec.~\ref{sec:method_protocol} defines the rotation-controlled evaluation protocol, Sec.~\ref{sec:method_refine} describes the Gravity-Aware Refinement (GAR) step, and Sec.~\ref{sec:method_metrics} defines the metrics used to trace errors across MDE point map, canonical meshes, layout, and physical plausibility.
\subsection{Rotation-Controlled Evaluation Protocol}
\label{sec:method_protocol}

\begin{figure*}[t]
\centering
\includegraphics[width=0.9\linewidth]{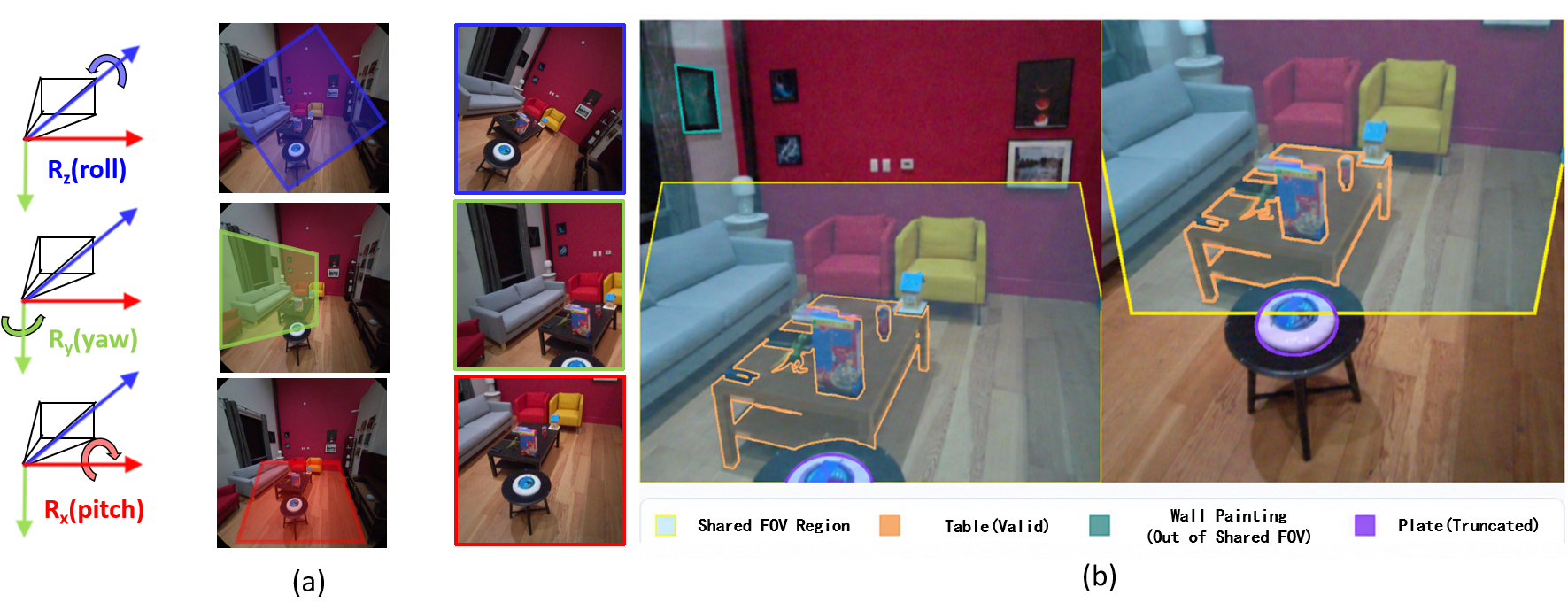}
\vspace{-0.1in}
\caption{In (a), rotated views are generated by camera-centric pure-rotation homographies along three axes from wide-FOV ADT images and cropped to a fixed FOV.
In (b), we evaluate only shared-FOV, non-truncated object instances.}
\label{fig:method_protocol}
\vspace{-0.2in}
\end{figure*}

As illustrated in Fig.~\ref{fig:method_protocol} (a), we construct rotated views from the original wide-FOV ADT images using camera-centric pure-rotation homographies. The original ADT view is treated as the reference view. We then sample single-axis rotations along roll, pitch, and yaw, represented by
\begin{equation}
 a=(\xi,\theta_a),
 \qquad
 \xi\in\{r,p,y\},
\label{eq:signed_rotation_set}
\end{equation}
where $\theta_a$ is the camera-rotation angle. We use this sweep to evaluate whether the same object remains consistent under controlled camera rotation. Since the transformation is a pure rotation around the optical center, it does not introduce camera translation, parallax, or newly observed object surfaces; the comparison therefore focuses on rotation-induced changes.

Let $R_a=R_{\xi}(\theta_a)\in\mathrm{SO}(3)$ denote the rotation that maps rays from the reference camera frame to the rotated camera frame. Let $\Omega_t$ be the fixed target crop domain, $\Pi_s$ the projection model of the original wide-FOV source image, and $\Pi_t^{-1}$ the back-projection model of the fixed target camera. For a target pixel $\mathbf u\in\Omega_t$ with homogeneous coordinate $\tilde{\mathbf u}=(u,v,1)^\top$, the source sampling location is obtained by inverse warping:
\begin{equation}
\mathbf r_t(\mathbf u)
=
\frac{\Pi_t^{-1}(\tilde{\mathbf u})}
{\left\|\Pi_t^{-1}(\tilde{\mathbf u})\right\|_2},
\qquad
\mathbf u_s(\mathbf u;a)
=
\Pi_s\!\left(R_a^{\top}\mathbf r_t(\mathbf u)\right).
\label{eq:ray_inverse_homography}
\end{equation}
The rotated view is obtained by sampling the original source image within the fixed target crop:
\begin{equation}
I_f^{(a)}(\mathbf u)
=
I_f^{(\mathrm{src})}\!\left(\mathbf u_s(\mathbf u;a)\right),
\qquad
\mathbf u\in\Omega_t .
\label{eq:constfov_crop_view}
\end{equation}
In the pinhole case with source and target intrinsics $K_s$ and $K_t$, Eq.~\eqref{eq:ray_inverse_homography} reduces to the standard pure-rotation homography
\begin{equation}
\tilde{\mathbf u}_s
\sim
H_{s\leftarrow a}\tilde{\mathbf u},
\qquad
H_{s\leftarrow a}=K_s R_a^{\top}K_t^{-1}.
\label{eq:pure_rotation_homography}
\end{equation}
Unlike padding-based rotation augmentation\cite{yang20253drot}, our construction keeps $\Omega_t$, $K_t$, and the output resolution fixed for all $a$. Thus each rotated view is a constant-FOV crop sampled from the original wide-FOV ADT image.

To avoid attributing visibility-induced errors to camera rotation, we apply the visibility-aware filtering shown in Fig.~\ref{fig:method_protocol}(b). An object instance is included only if its mask remains inside the shared FOV between the reference and rotated views and is not truncated by the image boundary in either view. We use each valid object mask as an evaluation unit, compute paired object-level measurements between the reference view and its rotated counterpart, and aggregate these measurements over frames, objects, axes, and rotation angles to obtain the final statistics.

\subsection{Gravity-Aware Refinement (GAR)}
\label{sec:method_refine}

Gravity-Aware Refinement (GAR) is a lightweight post-hoc refinement that stabilizes object rotations in the reconstructed layout. It is motivated by the empirical observation that canonical object meshes are more stable than camera-space layouts under camera rotation. We therefore treat the canonical object-frame structure as a weak prior. For tabletop and man-made objects, we assume the canonical $Z$ axis can be interpreted as the object's upright direction. When such objects are placed normally in a scene, their upright directions should be consistent with a shared scene gravity direction.

The gravity direction can be estimated from a single image by methods such as GeoCalib~\cite{veicht2024geocalib}. In this controlled study, we use the GT gravity instead of introducing an additional gravity estimator. For the rotated ADT views, this gravity cue is obtained from the reference-view gravity and the known camera rotation: $\mathbf g_{\mathrm{GT}}^{(a)}=R_a \mathbf g_{\mathrm{GT}}^{(0)}$, where $\mathbf g_{\mathrm{GT}}^{(0)}$ is the reference-view gravity direction and $R_a$ follows the rotation convention in Sec.~\ref{sec:method_protocol}. This keeps the refinement aligned with the controlled protocol and avoids adding another source of viewpoint-dependent noise.

For each object instance $i$, let $C_i$ be the canonical mesh points predicted by SAM3D, and $(R_i,t_i,s_i)$ be its initial camera-space layout. GAR only refines the object rotation to $\widehat R_i$, while keeping the predicted translation $t_i$ and scale $s_i$ fixed: $\widehat X_i=s_i(C_i\widehat R_i)+t_i$.

The refined rotation is optimized with two objectives. First, the refined object point cloud $\widehat X_i$ should stay aligned with the masked point-map segment $Y_i$ from the current input. Second, the refined object upright direction $\widehat{\mathbf u}_i$, obtained from the canonical $Z$ axis after applying $\widehat R_i$, should be parallel to the shared gravity cue. The loss is:
\begin{equation}
\mathcal L_{\mathrm{GAR}}
=
\frac{1}{|\mathcal I|}
\sum_{i\in\mathcal I}
\mathrm{CD}(\widehat X_i,Y_i)
+
\lambda_{\mathrm{grav}}
\frac{1}{|\mathcal I|}
\sum_{i\in\mathcal I}
\left(1-(\widehat{\mathbf u}_i^\top \mathbf g_{\mathrm{GT}}^{(a)})^2\right),
\label{eq:our_method_loss}
\end{equation}
where $\mathcal I$ is the valid object-instance set and $\mathrm{CD}$ denotes Chamfer distance. The first term keeps the refined object point cloud close to the observed point map, while the second term encourages all object upright axes to agree with the same gravity direction.

GAR should therefore be understood as a rotation-only stabilization step. It does not re-estimate object translation or scale, and it is not a full pose-and-scale optimization.

\subsection{Evaluation Metrics}
\label{sec:method_metrics}

We evaluate the pipeline at four levels: object point clouds from MDE, canonical meshes, layout, and physical plausibility. 

\subsubsection{Object Point Cloud}
\label{sec:method_metrics_mde}

We use object point cloud to denote the unordered set of 3D points extracted from the image-aligned MDE point map within an object mask.

Our main object-sensitive MDE metric is self-consistency normalized intra-object distortion (SC-nIoD). It evaluates whether the same object's point cloud remains structurally consistent when only camera rotation changes. For object instance $i$, let $P_i^{(0)}$ and $P_i^{(a)}$ be the mask-aligned object point clouds extracted from the reference and rotated MDE point maps. After the best cross-view similarity alignment, we define
\begin{equation}
\mathrm{SC\mbox{-}nIoD}(i,a)
=
\frac{
\mathrm{CD}\!\left(\mathcal{A}\!\left(P_i^{(a)}\right), P_i^{(0)}\right)
}{
L_{\mathrm{diag}}\!\left(P_i^{(0)}\right)
},
\label{eq:scniod}
\end{equation}
where $\mathcal{A}(\cdot)$ is the best cross-view Sim(3) alignment, $\mathrm{CD}$ is symmetric Chamfer distance, and $L_{\mathrm{diag}}(P_i^{(0)})$ denotes the 3D bounding-box diagonal length of the reference object point cloud. SC-nIoD is a unitless self-consistency discrepancy normalized by the reference object scale; a larger value means the predicted object geometry changes more across views.

We also report traditional scene-level MDE metrics such as AbsRel and normal-based depth error. Because these metrics are averaged over the scene, they may not reflect how the object point cloud itself changes under camera rotation; we therefore compare them with SC-nIoD.

\subsubsection{Canonical Mesh}
\label{sec:method_metrics_canonical}

We measure object-space canonical mesh predictions across views. Let $C_i^{(0)}$ and $C_i^{(a)}$ be the canonical point clouds sampled from the canonical meshes reconstructed from the reference and rotated views. After aligning the two canonical point clouds with ICP, we measure the post-alignment rotation error:
\begin{equation}
\mathrm{Rot}_{\mathrm{can}}(i,a)
=
\angle\!\left(R_{\mathrm{ICP}}(C_i^{(0)}, C_i^{(a)})\right).
\label{eq:canonical_icp_rot}
\end{equation}
It quantifies canonical-space self-consistency: if the canonical shape and orientation are stable under camera rotation, the post-alignment rotation error should remain small.

\subsubsection{Layout}
\label{sec:method_metrics_layout}
Let $X_i^{(0)}$ and $X_i^{(a)}$ be the reconstructed camera-space point clouds of the same object instance from the reference and rotated views. Before cross-view comparison, we compensate the known camera rotation from Eq.~\eqref{eq:signed_rotation_set} and express the rotated-view reconstruction in the reference camera frame:
\begin{equation}
X_i^{(a\rightarrow 0)} = \{R_a^{\top}\mathbf x \mid \mathbf x\in X_i^{(a)}\}.
\label{eq:undo_known_rotation}
\end{equation}
The layout metrics compare $X_i^{(0)}$ with $X_i^{(a\rightarrow 0)}$.

\paragraph{Pose}
Our main downstream orientation metric is centered ICP rotation:
\begin{equation}
\mathrm{ICP\mbox{-}C}(i,a)
=
\angle\!\Bigl(
R_{\mathrm{ICP}}(\bar X_i^{(0)}, \bar X_i^{(a\rightarrow 0)})
\Bigr),
\label{eq:icpc}
\end{equation}
where $\bar X$ denotes centering the point cloud at the origin and $R_{\mathrm{ICP}}$ is the post-alignment rotation returned by ICP. Lower is better.

We analyze MDE-to-layout error propagation by correlating matched error channels for each object-rotation pair after undoing the known camera rotation in Eq.~\eqref{eq:undo_known_rotation}. The MDE-side errors are extracted from the Sim(3) alignment between $P_i^{(0)}$ and $P_i^{(a)}$ used in Eq.~\eqref{eq:scniod}, while the layout-side errors are extracted from the relative transform between $X_i^{(0)}$ and $X_i^{(a\rightarrow 0)}$. The candidate channels are
\begin{equation}
q\in\{t_y,t_z,r_{\mathrm{yaw}},\|\Delta r\|_2,\|\Delta t\|_2,s-1\}.
\label{eq:prop_channels}
\end{equation}
Translation channels are normalized by the reference object diagonal $L_{\mathrm{diag}}$, and rotation channels are computed from the logarithm map of the relative rotation. The signed scale bias $s-1$ is used only for correlation analysis; the main scale metric is the non-negative drift in Eq.~\eqref{eq:scale_ratio_drift}.

For each sweep direction $\xi\in\{r,p,y\}$ and channel $q$, let $\mathbf{m}_q^\xi$ and $\mathbf{l}_q^\xi$ collect the MDE-side and layout-side errors over the same valid pairs with $a=(\xi,\theta_a)$. We compute
\begin{equation}
r_q^\xi=\mathrm{corr}_{\mathrm{P}}(\mathbf{m}_q^\xi,\mathbf{l}_q^\xi),\quad
\rho_q^\xi=\mathrm{corr}_{\mathrm{S}}(\mathbf{m}_q^\xi,\mathbf{l}_q^\xi).
\label{eq:prop_corr}
\end{equation}
Table~\ref{tab:error_propagation_directional} reports the statistically supported and interpretable channel couplings for \texttt{MOGEdep-SAM3D}.

\paragraph{Scale}
Since object layout includes both pose and object scale, we also evaluate scale consistency under camera rotation. For an estimated object scale $s_i^{(a)}$ at rotation $a$, we use fractional scale drift from the reference view,
\begin{equation}
\mathrm{ScaleErr}(i,a)
=
\max\!\left(
\frac{s_i^{(a)}}{s_i^{(0)}},
\frac{s_i^{(0)}}{s_i^{(a)}}
\right)-1.
\label{eq:scale_ratio_drift}
\end{equation}
This is reported as a unitless ratio, with $0$ indicating no scale drift.

\subsubsection{Physical Plausibility}
\label{sec:method_metrics_physics}

We measure physical plausibility by checking inter-object mesh penetration in each reconstructed scene. For the reconstructed-object set $\mathcal{I}$, let
$\mathcal{P}=\{(i,j): i,j\in\mathcal{I}, i<j\}$ denote all valid object pairs. For each pair $(i,j)$, we compute a normalized penetration fraction $\phi_{ij}$ and report the pair-normalized penetration rate
\begin{equation}
\mathrm{PenRate}
=
\frac{1}{|\mathcal{P}|}
\sum_{(i,j)\in\mathcal{P}}
\mathbf{1}[\phi_{ij}>\tau_p].
\label{eq:penrate}
\end{equation}
Lower is better. In the experiments, we summarize the rotation-induced change as
$\Delta=\mathrm{PenRate}_{\mathrm{rot}}-\mathrm{PenRate}_{\mathrm{orig}}$.

\section{EXPERIMENTS}
\label{sec:experiments}

Following the pipeline introduced in Sec.~\ref{sec:method_setup}, we present experimental results along object point clouds from MDE, canonical meshes, camera-space layout, and physical plausibility. We further use real-robot experiments and external baselines to validate our conclusions.

\subsection{Experimental Setup}
\label{sec:exp_setup}

We evaluate MDE models\cite{wang2025moge,wang2025vggt,lin2025depthanything3,piccinelli2025unidepthv2,hu2025metric3dv2}, two pipelines of SAM3D and external single-view mesh reconstruction pipelines\cite{dogaru2025gen3dsr,huang2025midi,meng2026scenegen,Zhao2025DepR}, on cropped images of ADT dataset\cite{pan2023adt} under the controlled-rotation protocol described in Sec.~\ref{sec:method_protocol}. Specifically, we sample frames at 30 fps from 236 ADT sequences. Starting from the original $1408\times1408$ images with a pinhole focal length of $f=610.94$ and a $98.1^\circ$ field of view, we generate cropped frames at $1408\times1408$ resolution with $f=1300$ and a $56.9^\circ$ field of view. We sample roll rotations within $\pm30^\circ$ and pitch/yaw rotations within $\pm20^\circ$.

\subsection{Qualitative Case}
\label{sec:exp_qualitative_case}

\begin{figure}
    \vspace{0.2in}
    \centering
    \includegraphics[width=0.8\linewidth]{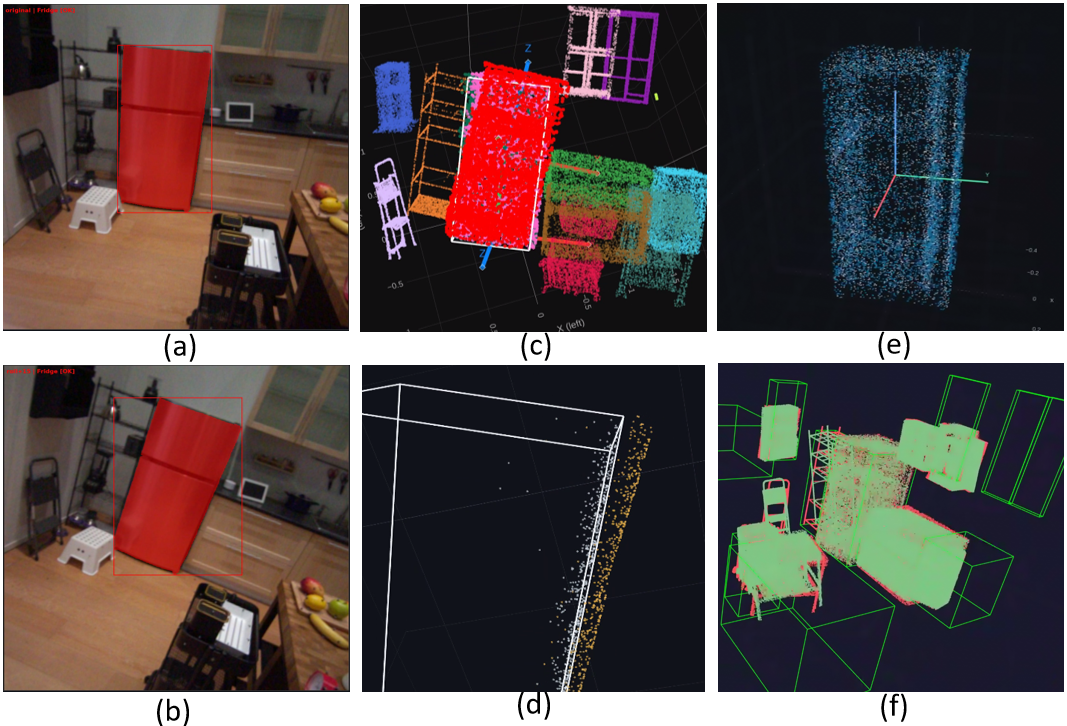}%
    \vspace{-0.1in}
    \caption{Qualitative case of a fridge in ADT\cite{pan2023adt} dataset. (a,b) show the original ADT upright view and the augmented roll view. For the fridge, (c) shows orientation inconsistency caused by camera rotation, (d) shows inconsistent distortion and surface normals in the MDE object point cloud, (e) shows that the canonical-space mesh remains well aligned after rotated, and (f) shows that our gravity-aware refinement (GAR) (red point cloud) corrects the erroneous rotation predicted by SAM3D (green point cloud).}
\label{fig:qualitative_case}
\end{figure}

Fig.~\ref{fig:qualitative_case} provides a qualitative example of a fridge in ADT. The MDE object point cloud and layout change under rotation, whereas the canonical mesh remains better aligned, and GAR corrects the orientation error.

\begin{figure}
\centering
\includegraphics[width=0.98\linewidth]{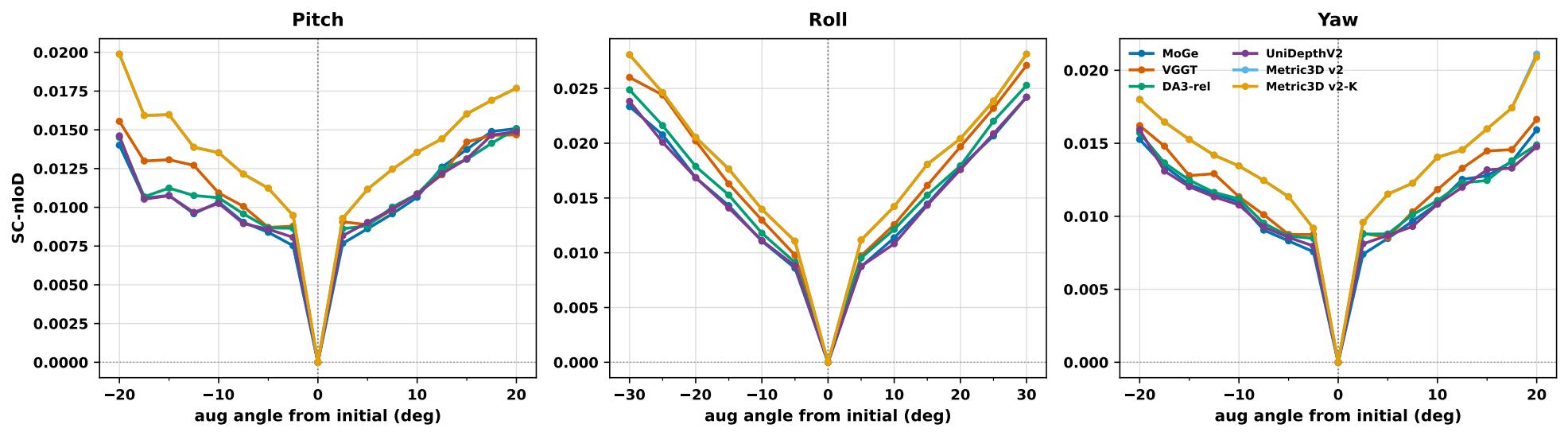}\\
\includegraphics[width=0.98\linewidth]{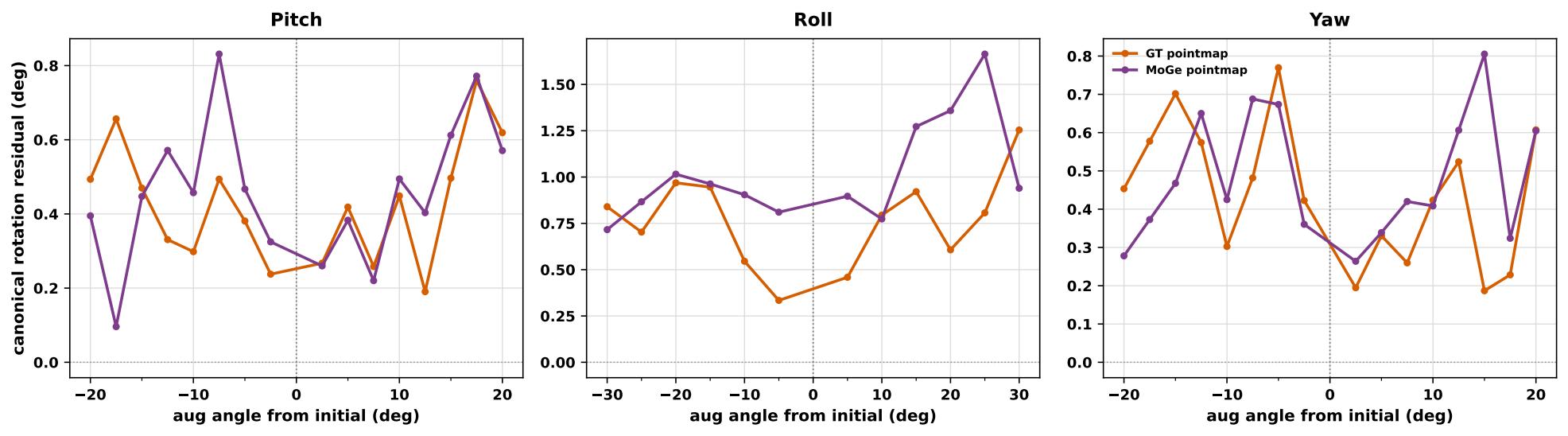}
\caption{Rotation effects on MDE object point cloud self-consistency (top) and canonical orientation stability (bottom). SC-nIoD shows a clear V-shaped response to camera rotation. Post-alignment canonical rotation errors remain relatively small.}
\label{fig:exp_mde_canonical}
\vspace{-0.20in}
\end{figure}

\subsection{Object Point Cloud from MDE}
\label{sec:exp_object}

We compare MoGe~\cite{wang2025moge}, VGGT~\cite{wang2025vggt}, DA3-rel~\cite{lin2025depthanything3}, UniDepthV2~\cite{piccinelli2025unidepthv2}, and Metric3D v2~\cite{hu2025metric3dv2}. We run VGGT in single-frame mode. Metric3D v2-K uses the calibrated camera intrinsics, while Metric3D v2 uses the model's fallback intrinsics when calibrated intrinsics are not supplied.

As shown in Fig.~\ref{fig:exp_mde_canonical} (top), object point clouds become less self-consistent as the rotation angle moves away from the reference orientation. Table~\ref{tab:exp_object} reports raw SC-nIoD for the object-level columns and scene-level deltas for AbsRel and NormErr, where $\Delta=\text{metric}(\text{rotated})-\text{metric}(\text{original})$. This distinction matters: the object-level self-consistency metric shows a stable rotation-dependent trend, whereas scene-level AbsRel and NormErr are mixed because they average over the full image. Thus, the MDE result is mainly used as evidence that object geometry changes with camera rotation, while the scene metrics serve as standard baselines.

\begin{table}
\vspace{0.2in}
\centering
\caption{Object-level MDE errors. SC-nIoD columns are raw axis-pooled medians over rotated views. Scene-level deltas are axis-matched means, with $\Delta=\mathrm{rotated}-\mathrm{original}$.}
\label{tab:exp_object}
\scriptsize
\setlength{\tabcolsep}{2.0pt}
\resizebox{0.98\linewidth}{!}{%
\begin{tabular}{lccccccccc}
\toprule
\multirow{2}{*}{\textbf{MDE model}} & \multicolumn{3}{c}{\textbf{Object self-consistency}} & \multicolumn{3}{c}{\textbf{Scene AbsRel delta}} & \multicolumn{3}{c}{\textbf{Scene NormErr delta (deg)}} \\
\cmidrule(lr){2-4}\cmidrule(lr){5-7}\cmidrule(lr){8-10}
 & \textbf{SC-nIoD$_p\downarrow$} & \textbf{SC-nIoD$_r\downarrow$} & \textbf{SC-nIoD$_y\downarrow$}
 & $\Delta_p\downarrow$ & $\Delta_r\downarrow$ & $\Delta_y\downarrow$
 & $\Delta_p\downarrow$ & $\Delta_r\downarrow$ & $\Delta_y\downarrow$ \\
\midrule
MoGe~\cite{wang2025moge} & \textbf{0.0099} & \textbf{0.0160} & \textbf{0.0104} & -0.013 & 0.008 & 0.001 & 0.759 & 6.139 & 2.087 \\
VGGT~\cite{wang2025vggt} & 0.0111 & 0.0181 & 0.0115 & \textbf{-0.024} & \textbf{-0.007} & -0.003 & 1.094 & \textbf{4.328} & 1.737 \\
DA3-rel~\cite{lin2025depthanything3} & 0.0106 & 0.0174 & 0.0109 & -0.019 & 0.005 & -0.001 & \textbf{0.668} & 6.006 & 2.016 \\
UniDepthV2~\cite{piccinelli2025unidepthv2} & 0.0102 & 0.0160 & 0.0105 & -0.008 & 0.006 & -0.002 & 1.513 & 6.195 & 1.479 \\
Metric3D v2~\cite{hu2025metric3dv2} & 0.0137 & 0.0201 & 0.0141 & -0.011 & 0.006 & \textbf{-0.004} & 1.359 & 5.916 & \textbf{0.916} \\
Metric3D v2-K~\cite{hu2025metric3dv2} & 0.0137 & 0.0201 & 0.0140 & -0.011 & 0.006 & \textbf{-0.004} & 1.359 & 5.916 & \textbf{0.916} \\
\bottomrule
\end{tabular}
}
\vspace{-0.2in}
\end{table}

\subsection{Object Canonical Mesh}
\label{sec:exp_canonical}

Fig. \ref{fig:exp_mde_canonical} (bottom) shows the opposite pattern from the plots of MDE-derived object point clouds: canonical meshes remain self-consistent across camera rotations. The post-alignment canonical rotation errors are all below two degrees, far smaller than the layout errors in Table~\ref{tab:exp_layout_physics_merged}.

\subsection{Layout and Physical Plausibility}
\label{sec:exp_layout}
\label{sec:exp_physics}

We use layout to mean camera-space object placement, including pose, translation, and scale. For downstream layout, we study both GT point map and MDE point map conditions, compare the one-stage pipeline against the two-stage pipeline, and then add Gravity-Aware Refinement (GAR). 

\begin{figure}[t]
\centering
\includegraphics[width=0.98\linewidth]{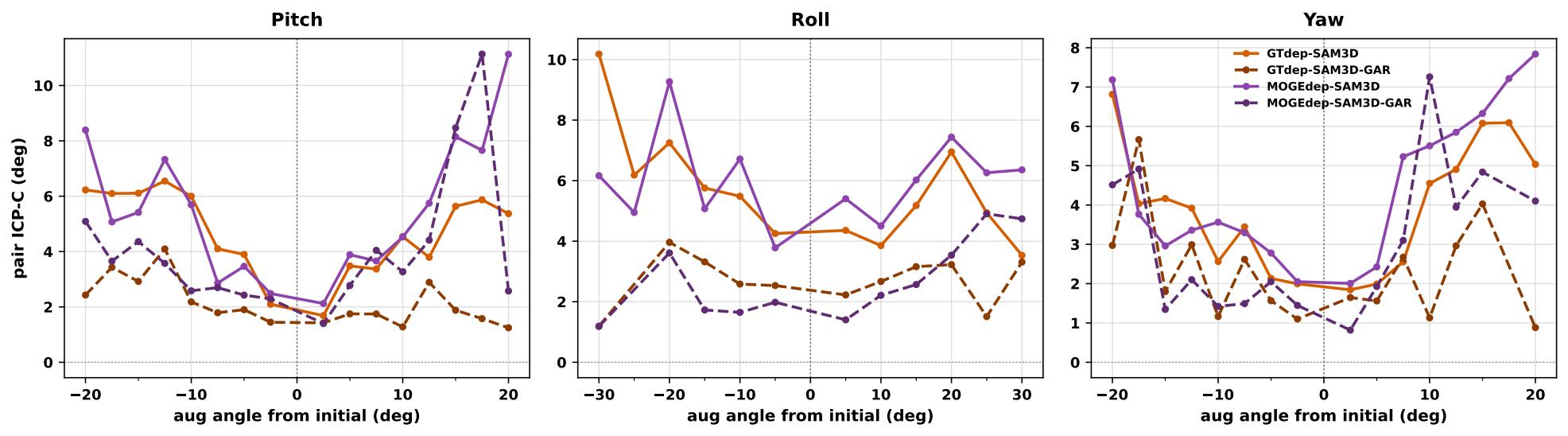}\\
\includegraphics[width=0.98\linewidth]{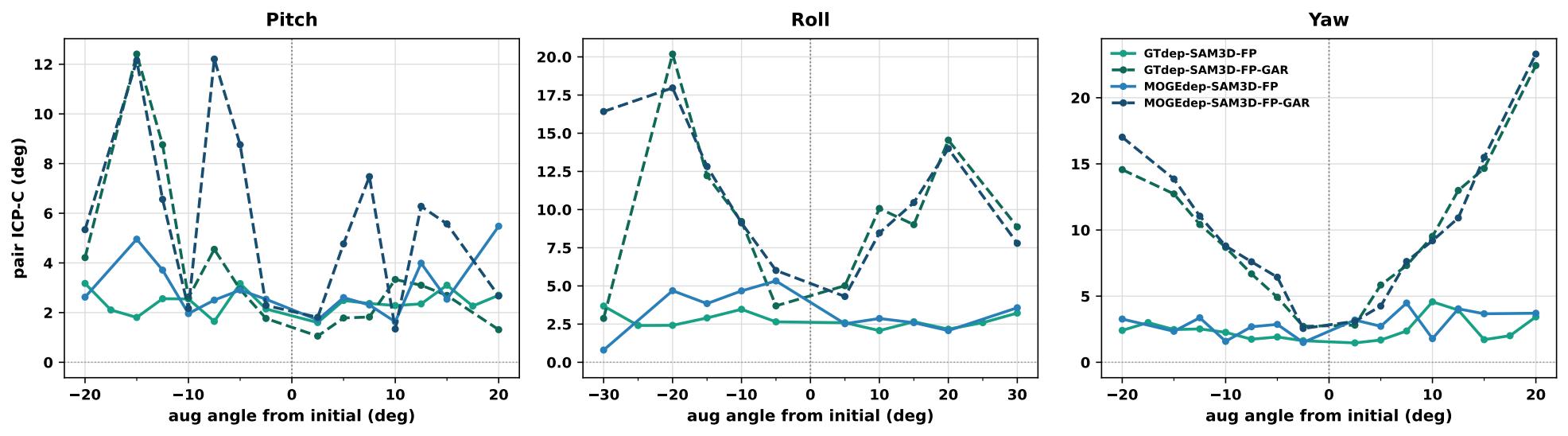}
\vspace{-0.1in}
\caption{One-stage (a) and two-stage (b) pairwise ICP-C under camera rotation. Our GAR improves the pairwise orientation consistency of the One-stage Pipeline. }
\label{fig:exp_layout}
\vspace{-0.1in}
\end{figure}

Fig. \ref{fig:exp_layout} is the main downstream result. The top row compares the one-stage pipeline, while the bottom row compares the two-stage pipeline with FoundationPose. The two-stage decomposition remains the clearest system result in the paper. Across pitch, roll, and yaw, \texttt{GTdep-SAM3D-FP} is the most stable pipeline family among the compared methods, while \texttt{MOGEdep-SAM3D} remains the most fragile. Table~\ref{tab:exp_layout_physics_merged} makes the same point in pooled form. \texttt{GTdep-SAM3D-FP} gives the lowest pairwise ICP-C ($2.404^\circ$) and the lowest normalized translation drift ($0.0198$). By contrast, \texttt{MOGEdep-SAM3D} remains both less stable and less accurate in camera-space layout, with pairwise ICP-C $4.731^\circ$ and normalized translation drift $0.0831$.

Gravity-Aware Refinement (GAR) improves the two one-stage pipeline variants in pairwise consistency: pairwise ICP-C drops from $4.148^\circ$ to $2.195^\circ$ for \texttt{GTdep-SAM3D}, and from $4.731^\circ$ to $2.414^\circ$ for \texttt{MOGEdep-SAM3D}. However, it degrades the two-stage pipeline variants: pairwise ICP-C rises from $2.404^\circ$ to $5.277^\circ$ for \texttt{GTdep-SAM3D-FP}, and from $2.902^\circ$ to $7.457^\circ$ for \texttt{MOGEdep-SAM3D-FP}. 

The PenRate column in Table~\ref{tab:exp_layout_physics_merged} measures how often reconstructed object pairs penetrate each other after rotation relative to the original views. Physical plausibility and geometric consistency do not collapse to a single ranking. For example, \texttt{MOGEdep-SAM3D-GAR} keeps its PenRate increase small ($0.005$), but its normalized ADT translation drift remains high ($0.0754$). This trade-off is consistent with GAR's intended scope: it stabilizes part of the orientation error, but it is not a full layout optimization.

\begin{table*}[t]
\vspace{0.1in}
\centering
\caption{Layout consistency in ADT and real-robot evaluations and physical plausibility. ADT translation is normalized by object scale. Real-robot translation reports absolute meter drift and normalized drift. Scale is the fractional drift in Eq.~\ref{eq:scale_ratio_drift}.}
\vspace{-0.1in}
\label{tab:exp_layout_physics_merged}
\scriptsize
\setlength{\tabcolsep}{2.0pt}
\resizebox{0.85\linewidth}{!}{%
\begin{tabular}{llrrrrrcr}
\toprule
\textbf{Pipeline} & \textbf{Method} 
& \multicolumn{3}{c}{\textbf{ADT layout}} 
& \multicolumn{1}{c}{\textbf{PenRate}} 
& \multicolumn{3}{c}{\textbf{Real robot layout}} \\
\cmidrule(lr){3-5}\cmidrule(lr){6-6}\cmidrule(lr){7-9}
& & Rot. $\downarrow$ (deg) & Trans. $\downarrow$ (ratio) & Scale $\downarrow$ (ratio)
& $\Delta\downarrow$ (ratio)
& Rot. $\downarrow$ (deg) & Abs / Norm Trans. $\downarrow$ (m / ratio) & Scale $\downarrow$ (ratio) \\
\midrule

\multirow{4}{*}{\shortstack{One-Stage\\Pipeline}}
& \texttt{GTdep-SAM3D} 
& 4.148 & \textbf{0.0294} & \textbf{0.0232} 
& \textbf{-0.010} 
& 8.167 & 0.0401 / 0.1956 & 0.0190 \\

& \texttt{GTdep-SAM3D-GAR} 
& \textbf{2.195} & 0.0354 & \textbf{0.0232} 
& 0.045 
& \textbf{6.598} & \textbf{0.0384 / 0.1871} & \textbf{0.0190} \\

& \texttt{MOGEdep-SAM3D} 
& 4.731 & 0.0831 & 0.0639 
& 0.064 
& 8.149 & 0.1385 / 0.2319 & 0.0559 \\

& \texttt{MOGEdep-SAM3D-GAR} 
& 2.414 & 0.0754 & 0.0639 
& 0.005 
& 6.702 & 0.1269 / 0.2124 & 0.0559 \\

\midrule

\multirow{4}{*}{\shortstack{Two-Stage\\Pipeline\\w/ FP}}
& \texttt{GTdep-SAM3D-FP} 
& \textbf{2.404} & \textbf{0.0198} & \textbf{0.0232} 
& 0.192 
& 5.952 & \textbf{0.0426 / 0.1890} & 0.0190 \\

& \texttt{GTdep-SAM3D-FP-GAR} 
& 5.277 & 0.0375 & \textbf{0.0232} 
& 0.079 
& \textbf{5.941} & 0.0428 / 0.1901 & \textbf{0.0190} \\

& \texttt{MOGEdep-SAM3D-FP} 
& 2.902 & 0.0411 & 0.0639 
& \textbf{0.052}
& 6.409 & 0.0428 / 0.2002 & 0.0559 \\

& \texttt{MOGEdep-SAM3D-FP-GAR} 
& 7.457 & 0.0642 & 0.0639 
& 0.099 
& 6.525 & 0.0447 / 0.2090 & 0.0559 \\

\bottomrule
\end{tabular}%
}
\vspace{-0.15in}
\end{table*}

\subsection{Error Propagation}
\label{sec:exp_propagation}

Table~\ref{tab:exp_layout_physics_merged} first shows that replacing GT point maps with MoGe point maps worsens downstream layout consistency. In the one-stage pipeline, rotation error increases from $4.148^\circ$ to $4.731^\circ$, normalized translation drift increases from $0.0294$ to $0.0831$, and scale drift increases from $0.0232$ to $0.0639$. The same trend appears in the two-stage pipeline with FoundationPose, where \texttt{MOGEdep-SAM3D-FP} is worse than \texttt{GTdep-SAM3D-FP} on all three ADT layout metrics.

To examine which error channels are associated with this degradation, Table~\ref{tab:error_propagation_directional} reports channel-level correlations for the primary MDE layout condition, \texttt{MOGEdep-SAM3D}. Each row correlates an MDE error channel with the matched downstream layout error channel over the same object-rotation pairs. The strongest supported associations appear in pitch-related depth/scale channels and roll-related orientation channels, suggesting that MDE errors are not transferred to layout as a single scalar error, but through axis-dependent channels.

\begin{table}[t]
\centering
\caption{Direction-resolved correlations between MoGe MDE error channels and \texttt{MOGEdep-SAM3D} layout error channels. $^*$, $^{**}$, and $^{***}$ denote $p<0.05$, $p<0.01$, and $p<0.001$, respectively.}
\label{tab:error_propagation_directional}
\scriptsize
\setlength{\tabcolsep}{2.0pt}
\newcommand{\sig}[1]{\makebox[1.4em][l]{\textsuperscript{#1}}}
\resizebox{0.75\linewidth}{!}{%
\begin{tabular}{llrr}
\toprule
Sweep & Matched error channel & Pearson $r$ & Spearman $\rho$ \\
\midrule
pitch & depth translation $t_z$ & 0.498\sig{*} & 0.336\sig{} \\
pitch & scale bias & -0.590\sig{**} & -0.213\sig{} \\
roll & yaw error $r_{\mathrm{yaw}}$ & -0.132\sig{} & -0.664\sig{***} \\
roll & rotation magnitude $\|\Delta r\|_2$ & 0.645\sig{***} & 0.296\sig{} \\
pitch & vertical translation $t_y$ & -0.313\sig{} & -0.705\sig{***} \\
roll & translation magnitude $\|\Delta t\|_2$ & -0.296\sig{} & -0.407\sig{*} \\
\bottomrule
\end{tabular}%
}
\vspace{-0.15in}
\end{table}

\subsection{Real-Robot Evaluation}
\label{sec:exp_real_robot}

\begin{figure}
\centering
\includegraphics[width=0.98\linewidth]{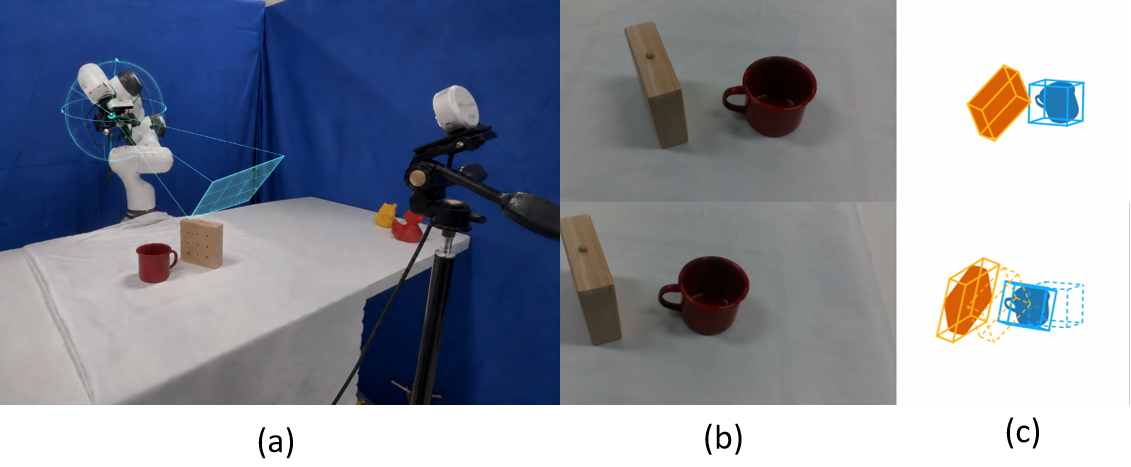}%
\vspace{-0.1in}
\caption{Experiment on the Franka wrist-mounted D435 camera. (a) The robot rotates the camera approximately around its optical center while maintaining object visibility.
(b) RGB observations before and after rotation.
(c) \texttt{GTdep-SAM3D} reconstruction results from the two views. Solid and dashed outlines denote the reconstructed object layouts before and after rotation, respectively. Their mismatch indicates translation and rotation inconsistency.}
\label{fig:real_robot}
\vspace{-0.10in}
\end{figure}

We further evaluate whether the same rotation sensitivity appears in a real-robot setup. As shown in Fig.~\ref{fig:real_robot}(a)(b), a Franka wrist-mounted D435 camera performs controlled rotational sweeps around the camera optical center while keeping the target objects visible.

For each frame, we run the same one-stage and two-stage pipelines. The real-robot results in Table~\ref{tab:exp_layout_physics_merged} follow the main ADT trend: the two-stage pipeline remains more stable layout. GAR improves one-stage rotation consistency. Fig.~\ref{fig:real_robot}(c) shows mismatched reconstructed layouts, indicating that camera rotation can still induce translation and rotation inconsistency in the real-robot setup.

\subsection{External Baselines}
\label{sec:exp_external_baselines}

\begin{table}[t]
\centering
\caption{External-baseline layout consistency}
\label{tab:external_baseline_layout_rts}
\scriptsize
\setlength{\tabcolsep}{3.5pt}
\resizebox{0.8\linewidth}{!}{%
\begin{tabular}{lrrr}
\toprule
\textbf{Method} & \textbf{Rot. $\downarrow$ (deg)} & \textbf{Trans. $\downarrow$ (ratio)} & \textbf{Scale $\downarrow$ (ratio)} \\
\midrule
\texttt{GTdep-SAM3D\cite{sam3d2025}} & 4.148 & 0.0294 & \textbf{0.0232} \\
\texttt{MIDI\cite{huang2025midi}} & \textbf{2.787} & 0.0824 & 0.0508 \\
\texttt{SceneGen\cite{meng2026scenegen}} & 6.088 & 0.1136 & 0.1019 \\
\texttt{DepR\cite{Zhao2025DepR}} & 4.212 & \textbf{0.0199} & 0.0990 \\
\texttt{Gen3DSR\cite{dogaru2025gen3dsr}} & 5.211 & 0.2937 & 0.0447 \\
\bottomrule
\end{tabular}%
}
\vspace{-0.20in}
\end{table}

Table~\ref{tab:external_baseline_layout_rts} extends the controlled-rotation comparison to external single-view mesh reconstruction baselines. The results are not a single-method ranking: \texttt{MIDI} has the lowest rotation error, \texttt{DepR} has the lowest translation drift, and \texttt{GTdep-SAM3D} gives the most stable scale. This supports the same multi-dimensional view of robustness used in our main evaluation: improving orientation alone does not guarantee stable camera-space translation or scale. 

\section{CONCLUSION}
\label{sec:conclusion}

We studied whether single-view mesh reconstruction can remain consistent under controlled robot camera rotation. Using roll, pitch, and yaw sweeps on ADT and a real Franka wrist-camera sequence, we traced rotation-induced errors across MDE point maps, canonical meshes, camera-space layouts, and physical plausibility. The results show that camera rotation is an important stressor for robot-facing single-view mesh reconstruction: it distorts monocular geometry, induces layout drift and penetration, and affects layout more strongly than canonical mesh prediction. The error propagation is also channel-specific, with selected MDE depth, scale, and orientation errors statistically coupled with downstream layout errors.

Our pipeline comparison further shows that architecture matters, but robustness is multi-dimensional. In the controlled-rotation setting, the two-stage SAM3D+FoundationPose pipeline is generally more stable than SAM3D's one-stage layout branch, while external baselines show that rotation, translation, and scale consistency do not always improve together. Gravity-Aware Refinement (GAR) improves one-stage orientation consistency by using gravity cues. This study is limited to isolated axis-wise rotations and a representative set of pipelines; future work should evaluate broader real-robot camera motions and develop reconstruction models that explicitly condition on camera rotation or gravity cues.

\bibliographystyle{IEEEtran}
\bibliography{bibliography/refs}

\end{document}